\documentclass{article}
\usepackage{spconf,amsmath,epsfig}

\let\OLDthebibliography\thebibliography
\renewcommand\thebibliography[1]{
  \OLDthebibliography{#1}
  \setlength{\parskip}{0pt}
  \setlength{\itemsep}{0pt plus 0.3ex}
}

\begin{document}\sloppy

\def\x{{\mathbf x}}
\def\L{{\cal L}}

\title{SC-ML: Self-supervised Counterfactual Metric Learning for \\
Debiased Visual Question Answering}
%
\name{Xinyao Shu\textsuperscript{1}, Shiyang Yan\textsuperscript{2}, Xu Yang\textsuperscript{3}, Ziheng Wu\textsuperscript{1}, Zhongfeng Chen\textsuperscript{1}, Zhenyu Lu\textsuperscript{1}$^{\ast}$ \thanks{*Corresponding author}}
%
\address{\textsuperscript{1}School of Artificial Intelligence, Nanjing University of Information Science and Technology \\
\textsuperscript{2}Inria, Universite Paris-Saclay\\
\textsuperscript{3}School of Computer Science and Engineering, Southeast University}

\maketitle

\begin{abstract}
Visual question answering (VQA) is a critical multimodal task in which an agent must answer questions according to the visual cue. Unfortunately, language bias is a common problem in VQA, which refers to the model generating answers only by associating with the questions while ignoring the visual content, resulting in biased results. We tackle the language bias problem by proposing a self-supervised counterfactual metric learning (SC-ML) method to focus the image features better. SC-ML can adaptively select the question-relevant visual features to answer the question, reducing the negative influence of question-irrelevant visual features on inferring answers. In addition, question-irrelevant visual features can be seamlessly incorporated into counterfactual training schemes to further boost robustness. Extensive experiments have proved the effectiveness of our method with improved results on the VQA-CP dataset. Our code will be made publicly available.

\end{abstract}

\section{Introduction}
The research in computer vision, natural language processing, and multimodal processing has made remarkable progress recently. Among them, with the emergence of large-scale datasets~\cite{VQA1.0,VQA2.0,gqa}, visual question answering (VQA) has been widely researched. Even though, VQA is still very challenging as it combines computer vision and natural language processing techniques. Moreover, VQA can bring significant social impacts, such as medical VQA, assistive devices for the blind, surveillance video queries, etc.

\begin{figure}[ht] \centering  
    \centering
    \includegraphics[width=0.9\linewidth]{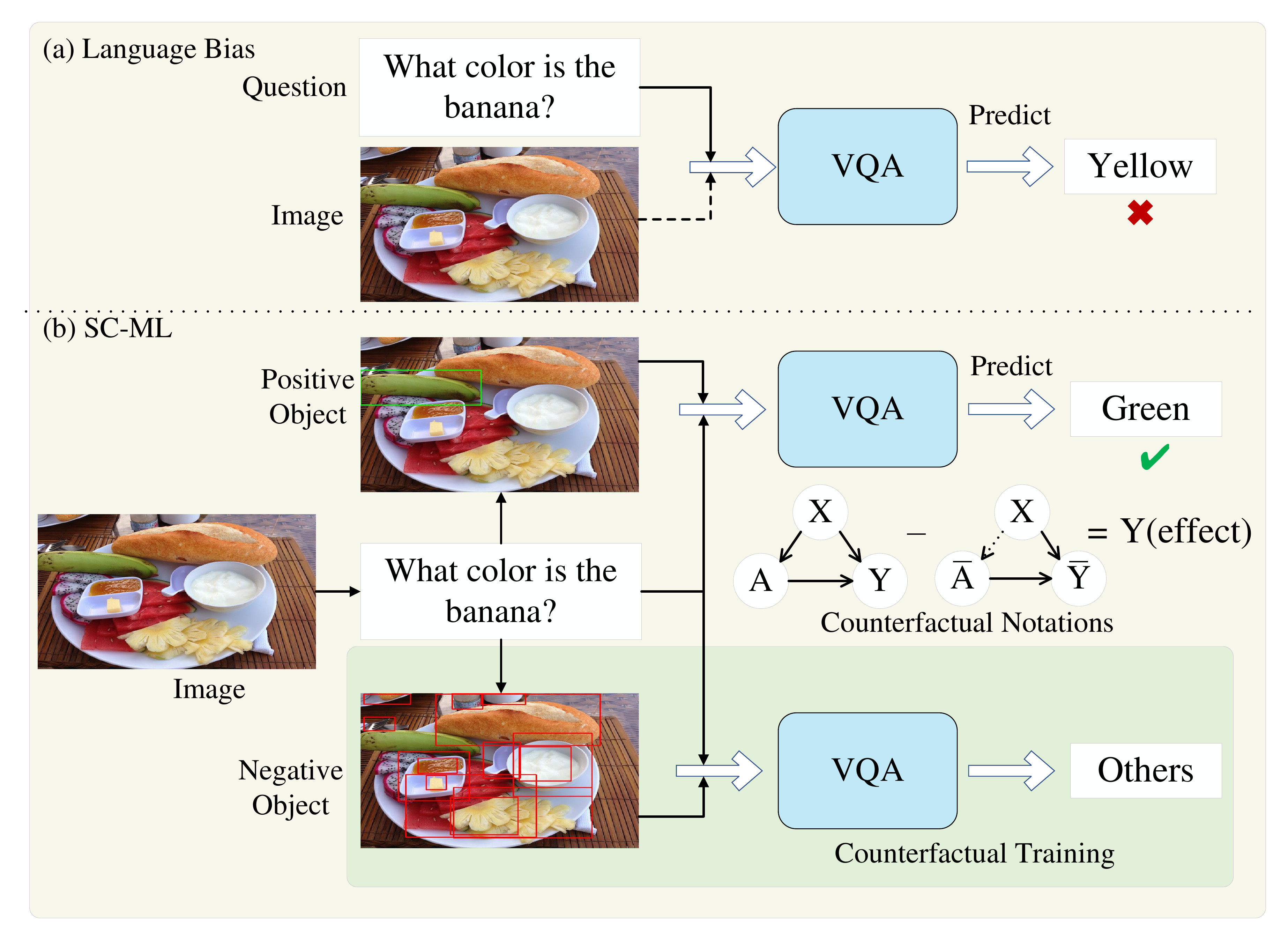}
    \caption{(a) Language bias: The VQA model generates answers based on the question without going through the image. (b) Our motivations: The model divides the image into question-relevant visual objects and question-irrelevant visual objects based on the question. The question-relevant visual objects are used to infer correct answers, and the question-irrelevant visual objects are used for counterfactual training. }
    \label{fig:introduction2}
\end{figure}

Visual question answering task requires the model to answer questions according to a given image. Both the image and the question are critical for the model to get the answer. However, recent research has shown that many of the current VQA models have language bias~\cite{VQA-CPv2}, i.e., the models only learn the associations between the questions and answers in the training set without making reasonable use of visual information. As shown in Fig.~\ref{fig:introduction2}(a), when faced with a problematic question-image pair, the model usually resorts to locking in the language prior knowledge in the training data and ignoring the images. Images are complex and rich, only a tiny region object (sometimes non-salient objects) helps answer the question. Therefore, during training, the model is more likely to reason directly based on the question alone or in combination with the background or salient objects in the image rather than based on the accurate relevant visual information, which causes the language bias problem. Language bias is very detrimental to the VQA model for practical applications in the real world. Because of language bias, the generalization ability and robustness of the model are limited~\cite{BUTD,lxmert,Ban}.


Most current VQA models use a conventional supervised learning method, i.e., the models are simply supervised by a final loss function without a powerful causal reasoning capability. This likelihood-based method only supervises the final predicted answer but ignores the true causal links between the question-image contents and the answer. For example, when gives a green banana image and asks ``What colour is the banana?'' (In Fig.~\ref{fig:introduction2}), the model is likely to predict the answer based on the question alone or in combination with the background (e.g. table or plate), ignoring the true cause (the association between "banana" and "yellow").

In this paper, to tackle the language bias and boost the generalization capacity of the VQA model, we propose a self-supervised counterfactual distance metric learning method. Specifically, as shown in Fig.~\ref{fig:introduction2}(b), we design a new self-supervised metric learning method. The method has an adaptive feature selection module that adaptively classifies visual features into question-relevant and question-irrelevant visual features. The VQA model inferences answer directly based on question-relevant visual features, ensuring the actual cause of the answer. Secondly, we construct counterfactual samples based on the question-irrelevant visual features to provide a counterfactual supervised signal for the model training without manual labelling, further reducing language bias.

In summary, our contributions are threefold: 1) We propose an adaptive self-supervised distance metric learning method that can focus image features adaptively to answer questions, ensuring the actual cause of the answers and thus alleviating language bias. 2) We propose a counterfactual training method that further encourages the model to learn more accurate visual attention to reduce language bias. 3) Comprehensive experimental results have validated the effectiveness of our approach, and we achieve state-of-the-art results on publicly available benchmark datasets.

\section{Related Work}

\subsection{Language Bias in VQA} Agrawal et al.~\cite{VQA-CPv2} pioneer the research of language bias. They reclassified the biased VQA dataset into the VQA-CP dataset, which worked on alleviating language bias from the dataset. Most current approaches to alleviating language bias can be broadly classified into four categories: 
Adding branch structure method~\cite{Areg,Rubi}, Answer-based method~\cite{VQA-CPv2,LMH,AdaVQA}, Data-balanced method~\cite{SSL,CSS,mutant} and Other method:
SAR~\cite{SAR} adopts answer re-ranking to address language biases. 
D-VQA~\cite{D-VQA} adopts two unimodal bias detection modules to recognize and remove the negative biases explicitly.

\subsection{Distance Metric Learning} 
Distance metric learning plays a significant role in a variety of computer vision applications, such as image retrieval~\cite{sohn2016improved}, cross-modal image-text matching~\cite{lee2018stacked}, person re-ID~\cite{hermans2017defense},  and transfer learning~\cite{oh2016deep}. Current research on distance metric learning focuses on the loss functions, e.g., Triplet loss~\cite{schroff2015facenet, hermans2017defense}, N-pair-mc~\cite{sohn2016improved}. There is also research work exploiting the mining techniques to consider the relationships between data samples, e.g., lifted structured~\cite{oh2016deep}, ranked list loss~\cite{wang2019ranked}, msloss~\cite{msloss}. 

\subsection{Self-supervised Learning} Self-supervised learning improves the feature extraction capability of the model by designing proxy tasks to mine the representational properties of the data itself as supervised information. CLIP~\cite{CLIP} uses contrastive self-supervised learning to learn multimodal representations of images and text. SSL~\cite{SSL} uses Self-Supervised Learning assisted tasks to help the model overcome language bias. In contrast to SSL, our efforts focus on learning self-supervised information from question-relevant visual information and counterfactual samples to alleviate language bias.

\subsection{Counterfactual Learning} Counterfactual learning has inspired several pieces of research in computer vision~\cite{cl2021}. Counterfactual learning has been exploited in recent VQA studies~\cite{SSL,CSS,CL,cl2020vqa}. In contrast to these efforts to generate counterfactual samples for debiased training, our efforts focus on adaptively mining the data samples for their self-biases and using these counterfactual information learning to improve the reasoning capability of the model.

\section{Proposed method}
\begin{figure*}[ht]
    \centering
    \includegraphics[width=0.8\linewidth]{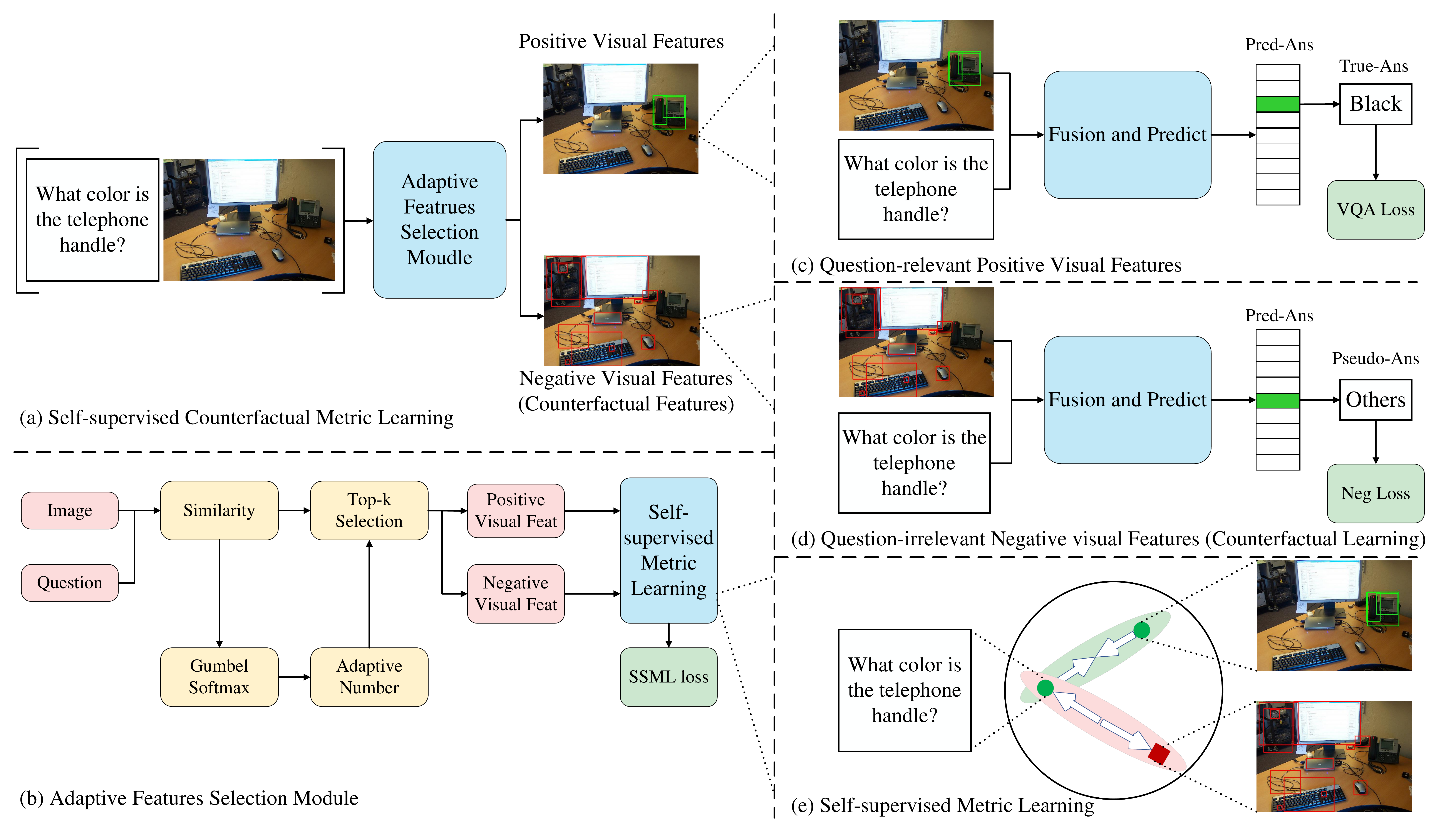}
    \caption{A schematic diagram of the proposed method: We propose the SC-ML for adaptive feature selection, dividing the image features into relevant and irrelevant ones. The relevant features are applied in VQA training, while the irrelevant features assist the model training via counterfactual reasoning.}
    \label{fig:model}
\end{figure*}


\subsection{Baseline Model} 
We adopt the LXMERT~\cite{lxmert} model as the baseline model in our research. LXMERT is a dual-stream Transformer architecture consisting of two unimodal encoders for visual and language modalities, respectively, and a cross-modal encoder for aligning entities across both modalities. In this paper, we split the cross-modal encoder of the LXMERT into two parts. The first part is loaded with pre-training parameters for multimodal alignment, and the second is self-defined and without pre-training parameters and employs ReLU~\cite{relu} as the activation function for multimodal fusion. 

\subsection{Feature Selection Module} 

For each image, the LXMERT uses an image encoder $e_v$ to output a set of visual region features: $V=\left[v_{1} ; \ldots \ldots ; v_{n}\right] \in \mathrm{R}^{n \times d}$, where $v_i$ is i-th object feature. For each question Q, the LXMERT uses a question encoder $e_q$ to output a set of word features: $Q=\left[q_{1} ; \ldots \ldots ; q_{m}\right] \in \mathrm{R}^{m \times d}$, where $q_j$ is j-th word.

Subsequently, we adopt cosine similarity to calculate the correlation between the visual region features $V$ and the question features $Q$. 
\begin{equation}
\begin{aligned}
{Sim}_{k}=\sum_{s=1}^{m} {Cosine}\left(V_{k}, Q_{s}\right), \\
k \in[1, n], s \in[1, m], 
\end{aligned}
\end{equation}
where ${Sim}_{k}$ denotes the similarity representation of the $k$-th visual feature to the question feature sequence, the similarity vector $Sim$ as a whole is represented as ${Sim}=\left[{Sim}_{1}, ; \ldots \ldots ; {Sim}_{n}\right]$. 

As shown in Fig.~\ref{fig:model}(a), We select the top $k$ visual features with higher $Sim$ as the question-relevant visual features and the remaining $n-k$ visual features as the question-irrelevant visual features. 
\begin{equation}
\begin{aligned}
positive & = \left\{\begin{array}{l}V[k],  \text{if} \ k  \ \text{in} \ index ({Top\text{-}k}({Sim})) \\ 0, { otherwise },\end{array}\right. 
\\
negative & = \left\{\begin{array}{l} 0,  \text{if}  \ k \  \text{in} \ index (Top\text{-}k({Sim})) \\
V[k] {, otherwise },
\end{array}\right.
\end{aligned}
\end{equation}
where positive denotes question-relevant visual features and negative denotes question-irrelevant visual features.

\subsection{Self-supervised Metric Learning Module} 
Though by the feature selection module, we have grouped the visual features according to the question, some visual features may still be near the decision boundary, causing an incorrect division. To avoid this, we adopt Multi-Similarity Loss (ms loss)~\cite{msloss} to learn an embedding space (as Fig.~\ref{fig:model}(e)) that makes the distance between similar samples closer and the distance between different samples far away. 

Compared with other metric learning loss, ms loss considers both the self-similarity with sample pairs and the relative similarity between sample pairs. It uses a weighting approach to obtain sample pairs with higher informativeness. 
The ms loss can be expressed in a specific way:
\begin{equation}\label{ms_eq}
\begin{aligned}
\mathcal{L}_{ms}=\frac{1}{m} \sum_{i=1}^{m} &\left\{\frac{1}{\alpha} \log \left[1+\sum_{k \in \mathcal{P}_{i}} e^{-\alpha\left(S_{i k}-\lambda\right)}\right]\right.\\
+&\left.\frac{1}{\beta} \log \left[1+\sum_{k \in \mathcal{N}_{i}} e^{\beta\left(S_{i k}-\lambda\right)}\right]\right\},
\end{aligned}
\end{equation}
where $S_{ik}$ is the cosine similarity of the sample pair, $\lambda$ is the similarity margin, $\alpha$ and $\beta$ are hyperparameters.

\subsection{Counterfactual Learning Module}
According to the feature selection module, we classify the visual features into positive and negative features based on the question. In the answer prediction module, we use only positive features to predict answers (as Fig.~\ref{fig:model}(c)). Our intuition is that negative features should not contain question-relevant information. To better learn the relationship between positive and negative features, we propose a counterfactual learning module that takes negative features as counterfactual features to train the model in order to improve the robustness of the model. Specifically, we fuse the counterfactual features with the question features and use the same answer prediction module to predict the answer (as Fig.~\ref{fig:model}(d)); we call the predicted answer $Pred_{neg}$. For the same reason, we call the answer predicted by the positive feature $Pred_{pos}$. As the counterfactual features do not contain features related to the question, $Pred_{neg}$ should get different answers from $Pred_{pos}$. We assign a pseudo-label to $Pred_{neg}$ that is different from the standard answer, denoted as $Ans_{pseudo}$. $Ans_{pseudo}$ is to remove the first $n$ answers predicted by $Pred_{pos}$ in the labeled answer $Ans$. If $Pred_{pos}$ is the same as $Ans$, then $Ans_{pseudo}$ is all zeros. It can be specifically expressed as:
\begin{equation}
\begin{aligned}
Ans_{pseudo}=\{Ans_{i} \mid Ans_{i} \in Ans, \\ Ans_{i} \notin Top\text{-}n({Pred}_{pos})\},
\end{aligned}
\end{equation}
where Ans denotes the answer set, $Ans_{i}$ denotes the i-th answer in the answer set, and $Top\text{-}n$ denotes the first top $n$ samples. 

Finally, the training loss of positive features and counterfactual features in the VQA task can be expressed as:
\begin{equation}
\begin{aligned}
\begin{array}{c}
 Loss_{VQA}\left(Pred, Ans\right)=-\frac{1}{N} \sum_{i=1}^{N} Ans_{i} \log \left(\sigma\left({Pred}_{i}\right)\right) \\
+\left(1-Ans_{i}\right) \log \left(1-\sigma\left({Pred}_{i}\right)\right), \\
 Loss_{pos}={Loss}_{VQA}\left( {Pred}_{pos}, Ans\right), \\
 Loss_{neg}={Loss}_{VQA}\left( {Pred}_{neg}, Ans_{pseudo}\right),
\end{array}
\end{aligned}
\end{equation}
Where ${Loss}_{VQA}$ is binary cross-entropy loss, $\sigma$ is the sigmoid activation function, $Loss_{pos}$ is the training loss of positive features, and $ Loss_{neg}$ is the training loss of counterfactual learning.

The overall loss function contains three parts:
\begin{equation}
\begin{aligned}
Loss={Loss}_{pos} + \gamma*({Loss}_{neg}+ {Loss}_{ms}),
\end{aligned}
\end{equation}
where ${Loss}_{pos}$ is the optimization scheme for predicting correct answers by positive features; ${Loss}_{neg}$ is counterfactual reasoning loss; ${Loss}_{ms}$ is the metric learning loss, and $\gamma$ is factor bwtween ${Loss}_{neg}$ and ${Loss}_{ms}$ .

\subsection{Adaptive Feature Selection Strategy} 
\begin{figure}[ht]
    \centering
    \includegraphics[width=0.8\linewidth]{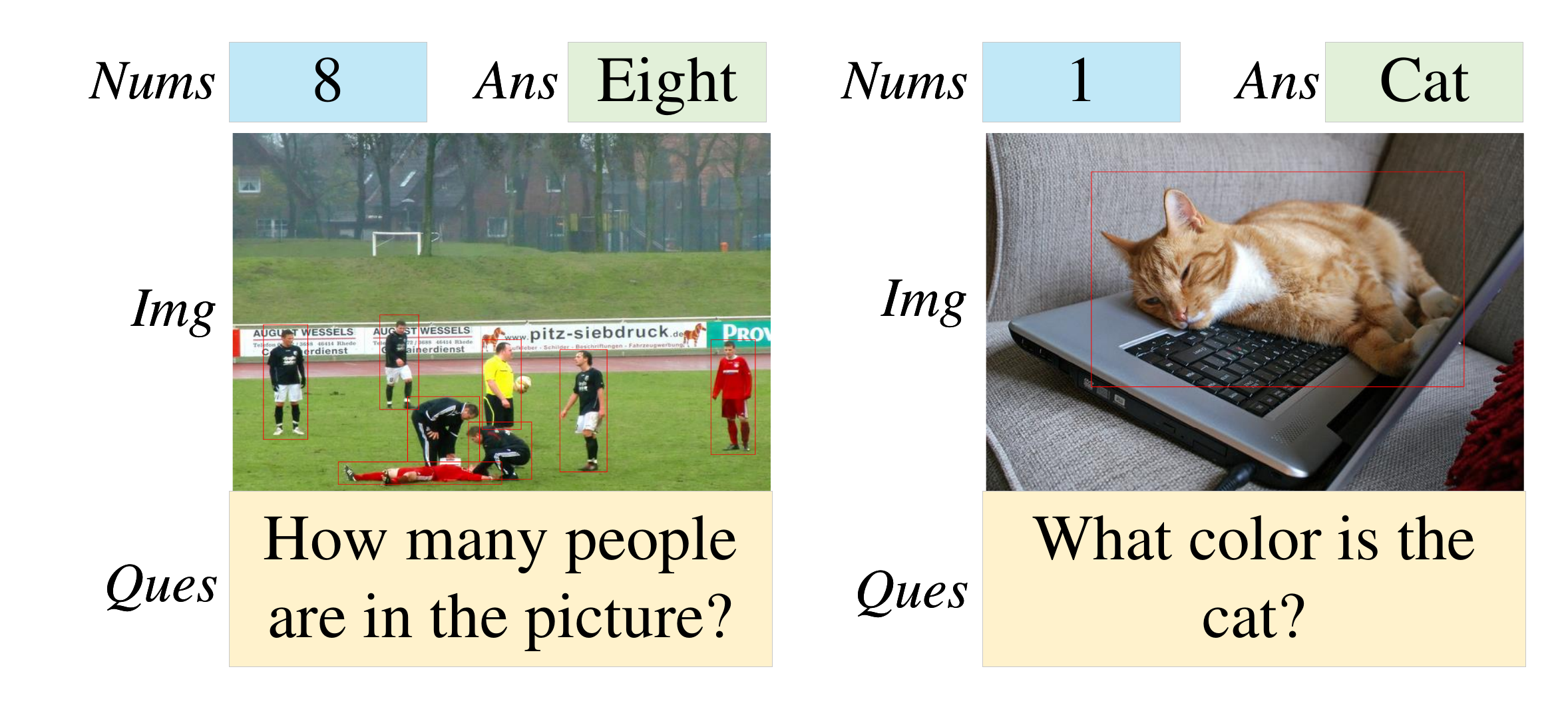}
    \caption{Different pairs of questions and images require different numbers of features to be selected for the image regions related to the question. }
    \label{fig:adatpive}
\end{figure}
Using a fixed selection of visual region features as positive features does not apply to all VQA questions. For example, a fixed selection of $k$ visual features as positive features will present particular problems. 
As shown in the left of Fig.~\ref{fig:adatpive}, when asked ``how many people are in the picture?'', at least $8$ visual region features of the person are needed to answer the question altogether. However, as shown in the right of Fig.~\ref{fig:adatpive}, when asked ``what colour is the cat?'', only $1$ feature of the cat can answer the question altogether.
Therefore, if a fixed selection strategy is adopted, $8$ features are selected as the number of the question-relevant visual features for all questions. Unfortunately, for the right figure, the remaining $7$ question-irrelevant visual features in Fig.~\ref{fig:adatpive} right may all be interference features, which still negatively impact the model's inference. If $1$ feature is selected as the fixed number of the question-relevant visual feature, the other $7$ question-relevant visual features in Fig.~\ref{fig:adatpive} will be used for counterfactual training, which is contrary to the fact.

Therefore, we propose an adaptive feature selection strategy, where we apply Gumbel-Softmax~\cite{gumbel} to output an index topology from a similarity vector (as Fig.~\ref{fig:model}(b)). 
Then we apply the masking operation to select the visual features relevant to the question. Finally, by applying Gumbel-Softmax and the masking operation, we realize an adaptive trainable Top-k operation. The proposed adaptive trainable Top-k operation avoids the laborious tuning of $k$ hyper-parameters in traditional schemes. It is a general algorithm that can be easily extended to many other applications. Specifically, the adaptive Top-k is implemented by a Gumbel-Softmax and a masking technique to achieve the back-propagation capability:
\begin{equation}
\begin{aligned}
\begin{array}{c}
 { k }= Gumbel\_softmax ( { Sim }), \\
 { mask }= { Ones }( {k}-1)+ { One\_hot }({k}), \\
 { positive }=V *  { mask }, \\
 { negative }=V *(1-{mask}),
\end{array}
\end{aligned}
\end{equation}
where $Ones(dim)$ denotes the generation of an all-1 vector, and $One\_hot$ means the one-hot embedding. The operator $+$ denotes the adding operation for vectors, and the operator $*$ is the element-wise product. We use Gumbel-Softmax to automatically generate Top-k values and integrate the $k$ value into the training of the entire model. We make the Top-k scheme trainable, in terms of the $k$, via a masking operation simply because all the operations involved are continuous.

\begin{figure*}[htbp]
\centering
\includegraphics[width=0.8\linewidth]{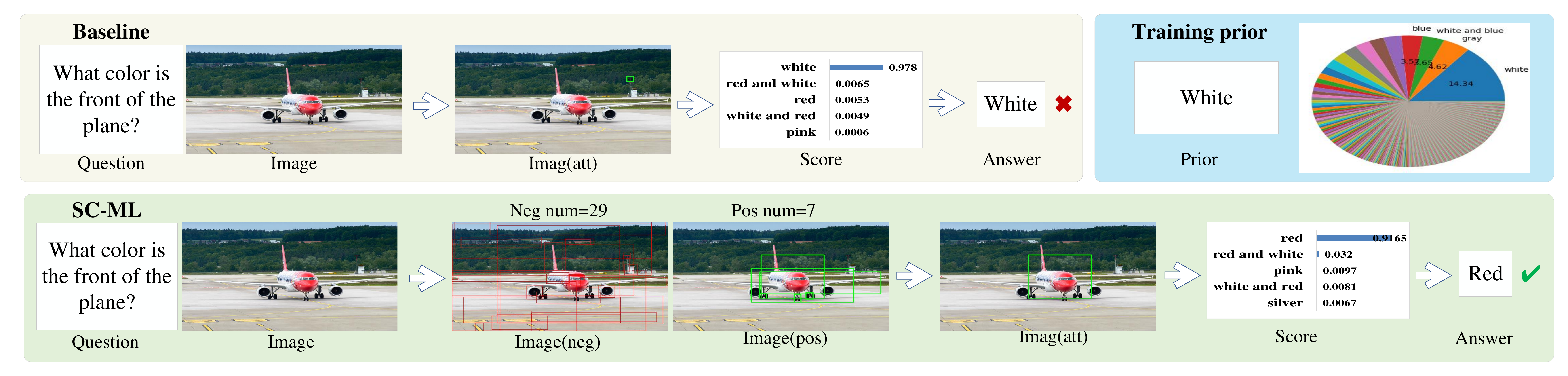}
\caption{Visualizations for mitigating the language bias. The blue area shows the language bias present in the current question.}
\label{fig:icme}
\end{figure*}

\section{Experiments}
\subsection{Implementation details}
\textbf{Datasets.} To validate the effectiveness of our method, we evaluate our method on the VQA CP~\cite{VQA-CPv2} dataset. The VQA CPv2 and VQA CPv1 datasets are two standard benchmarks for estimating the ability of models to overcome language bias problems in VQA, and they reorganize the VQA v1~\cite{VQA1.0} datasets and VQA v2~\cite{VQA2.0} so that the answers to each question category have different distributions in the training and test sets. We evaluate the model by adopting standard VQA evaluation metrics.  
\textbf{Training.} Our model is trained for 20 epochs with a batch size of 128, an optimizer of Adamax, and a learning rate of 5e-5. In ms loss (as Equation~\ref{ms_eq}), $\lambda$ is set to 0.5, $\alpha$ is set to 2 and $\beta$ is set to 50. 

\begin{table}[] \centering
\caption{Ablation study on the VQA CPv2 dataset.}
\label{tab:ablation_all}
\resizebox{\linewidth}{!}{
\begin{tabular}{lllll}
\hline
Methods & Overall & Yes/no & Number & Other \\ \hline
LMH-LXMERT & 58.51 & 53.38 & 62.00 & 60.25 \\
+ pos & 67.10 & 84.08 & 60.52 & 60.01 \\
+ pos + ms& 67.16 & 84.40 & 60.36 & 59.99 \\
+ pos + neg + ms & 67.20 & 83.08 & 64.80 & 59.54 \\
\textbf{+ pos + neg + ms + adaptive} & \textbf{68.42} & \textbf{87.57} & \textbf{63.07} & \textbf{59.86} \\ \hline
\end{tabular}}
\end{table}

\subsection{Ablation Study}
We provide SC-ML ablation experiments based on the VQA CPv2 dataset. We performed all the ablation experiments by building on top of the original LXMERT model~\cite{lxmert} and LMH~\cite{LMH} based on LXMERT model (LHM-LXMERT). \noindent
\textbf{Feature Selection Module.} From Table~\ref{tab:ablation_all}, `pos' means inferring answers by the question-relevant visual positive features selected in Feature Selection Module. The performance of masking negative visual features to infer answers using only positive visual features is improved over the performance of using all visual features to infer answers. It proves that there is significant redundancy in visual information that affects the model's reasoning about visual information, which leads to the model inferring answers based on questions only (language bias). It also validates the effectiveness of our method in alleviating language bias. \noindent
\textbf{Self-supervised Metric Learning. }From Table~\ref{tab:ablation_all}, `ms' denotes the Self-supervised Metric Learning module. When the Self-supervised Metric Learning Module is added to the model, there is some improvement in accuracy. Because Self-supervised Metric Learning helps to mine the relationship between question-relevant positive visual features and question-irrelevant negative visual features during model training.
\noindent
\textbf{Counterfactual Learning Module. }From Table~\ref{tab:ablation_all}, `neg' indicates the Counterfactual Learning Module. When the Self-supervised Learning Strategy is added, there is a certain improvement in model performance, which validates the effectiveness of the Counterfactual Learning Module on alleviating language bias. \noindent
\textbf{Adaptive Feature Selection Strategy. } From Table~\ref{tab:ablation_all}, `adaptive' indicates Adaptive Feature Selection Strategy. Other methods use a fixed selection of 15 features. More specific results are shown in supplementary materials. The fixed feature selection strategy is likely to lead to wrongly viewing question-relevant positive visual features as counterfactual features, which is contrary to the facts. `adaptive' indicates Adaptive Feature Selection Strategy, which is significantly better than other fixed feature selection methods. The adaptive Feature Selection Strategy avoids expensive tuning between different datasets and is suitable for generic scenarios.

\begin{table}[] \centering
\caption{The results on the VQA CPv2 dataset.}
\label{tab:cpv2}
\resizebox{0.95\linewidth}{!}{
\begin{tabular}{llcccc}
\hline
{Methods} & {Base} & Overall & Yes/No & Num & other \\ \hline
LMH~\cite{LMH} & BUTD & 52.45 & 69.81 & 44.46 & 45.54 \\
LMH-CSS~\cite{CSS} & BUTD & 58.95 & 84.37 & 49.42 & 48.21 \\
LMH-CSS-CL~\cite{CL} & BUTD & 59.18 & 86.99 & 49.89 & 47.16 \\
LMH~\cite{LMH} & LXMERT & 58.51 & 53.38 & 62.00 & 60.25 \\
LMH-CSS~\cite{CSS} & LXMERT & 63.63 & 84.70 & 62.12 & 53.00 \\
LMH-CSST~\cite{CSST} & LXMERT & 65.71 & 90.10 & 63.70 & 53.48\\
LMH-SAR~\cite{SAR} & LXMERT  & 66.73 & 86.00 & 62.34 & 57.84 \\ \hline
\textbf{LMH-Ours(SC-ML)} & \textbf{LXMERT}  & \textbf{68.42} & \textbf{87.57} & \textbf{63.07} & \textbf{59.86} \\ \hline
\end{tabular}}
\end{table}


\subsection{Comparison with State-of-the-arts}
We evaluated our method (SC-ML) on the VQA CPv2 benchmark (as shown in Table.~\ref{tab:cpv2}). Our method significantly outperforms previous methods and achieves state-of-the-art performance on the VQA CPv2. We have visualized the SC-ML method and compared it with the baseline (LXMERT) method. As shown in Fig.~\ref{fig:icme}, SC-ML can effectively alleviate language bias by adaptively masking the interfering visual features with counterfactual learning. We provide more results and visualizations in the supplementary.

\section{Conclusion}
This paper proposes a self-supervised counterfactual distance metric learning method (SC-ML) to alleviate the language bias problem in VQA tasks. SC-ML adaptively selects question-relevant image features to answer the questions effectively, alleviating language bias and improving model robustness. In addition, the negative features are seamlessly combined with counterfactual reasoning, further improving the final performance. Comprehensive experiments validate the ability of the Self-supervised Distance Metric Learning method to alleviate language bias and achieve state-of-the-art results on the VQA-CP dataset. 


\bibliographystyle{IEEEbib}
\bibliography{icme2023template}

\begin{thebibliography}{10}

\bibitem{VQA1.0}
S.~Antol, A.~Agrawal, J.~Lu, M.~Mitchell, D.~Batra, C.~L. Zitnick, and
  D.~Parikh,
\newblock ``Vqa: Visual question answering,''
\newblock in {\em ICCV}, 2015.

\bibitem{VQA2.0}
Y.~Goyal, T.~Khot, D.~Summers-Stay, D.~Batra, and D.~Parikh,
\newblock ``Making the v in vqa matter: Elevating the role of image
  understanding in visual question answering,''
\newblock in {\em CVPR}, 2017.

\bibitem{gqa}
D.~A. Hudson and C.~D. Manning,
\newblock ``Gqa: A new dataset for real-world visual reasoning and
  compositional question answering,''
\newblock in {\em CVPR}, 2019.

\bibitem{VQA-CPv2}
A.~Agrawal, D.~Batra, D.~Parikh, and A.~Kembhavi,
\newblock ``Don't just assume; look and answer: Overcoming priors for visual
  question answering,''
\newblock in {\em CVPR}, 2018.

\bibitem{BUTD}
P.~Anderson, X.~He, C.~Buehler, D.~Teney, M.~Johnson, S.~Gould, and L.~Zhang,
\newblock ``Bottom-up and top-down attention for image captioning and visual
  question answering,''
\newblock in {\em CVPR}, 2018.

\bibitem{lxmert}
H.~Tan and M.~Bansal,
\newblock ``Lxmert: Learning cross-modality encoder representations from
  transformers,''
\newblock in {\em EMNLP}, 2019.

\bibitem{Ban}
J.~Kim, J.~Jun, and B.~Zhang,
\newblock ``Bilinear attention networks,''
\newblock in {\em NeurIPS}, 2018.

\bibitem{Areg}
S.~Ramakrishnan, A.~Agrawal, and S.~Lee,
\newblock ``Overcoming language priors in visual question answering with
  adversarial regularization,''
\newblock in {\em NeurIPS}, 2018.

\bibitem{Rubi}
R.~Cadene, C.~Dancette, M.~Cord, D.~Parikh, et~al.,
\newblock ``Rubi: Reducing unimodal biases for visual question answering,''
\newblock in {\em NeurIPS}, 2019.

\bibitem{LMH}
C.~Clark, M.~Yatskar, and L.~Zettlemoyer,
\newblock ``Don’t take the easy way out: Ensemble based methods for avoiding
  known dataset biases,''
\newblock in {\em EMNLP}, 2019.

\bibitem{AdaVQA}
Y.~Guo, L.~Nie, Z.~Cheng, F.~Ji, J.~Zhang, and A.~Del~Bimbo,
\newblock ``Adavqa: Overcoming language priors with adapted margin cosine
  loss,''
\newblock in {\em IJCAI}, 2021.

\bibitem{SSL}
X.~Zhu, Z.~Mao, C.~Liu, P.~Zhang, B.~Wang, and Y.~Zhang,
\newblock ``Overcoming language priors with self-supervised learning for visual
  question answering,''
\newblock in {\em IJCAI}, 2021.

\bibitem{CSS}
L.~Chen, X.~Yan, J.~Xiao, H.~Zhang, S.~Pu, and Y.~Zhuang,
\newblock ``Counterfactual samples synthesizing for robust visual question
  answering,''
\newblock in {\em CVPR}, 2020.

\bibitem{mutant}
T.~Gokhale, P.~Banerjee, C.~Baral, and Y.~Yang,
\newblock ``Mutant: A training paradigm for out-of-distribution generalization
  in visual question answering,''
\newblock in {\em EMNLP}, 2020.

\bibitem{SAR}
Q.~Si, Z.~Lin, M.~yu~Zheng, P.~Fu, and W.~Wang,
\newblock ``Check it again: Progressive visual question answering via visual
  entailment,''
\newblock in {\em ACL}, 2021.

\bibitem{D-VQA}
Z.~Wen, G.~Xu, M.~Tan, Q.~Wu, and Q.~Wu,
\newblock ``Debiased visual question answering from feature and sample
  perspectives,''
\newblock in {\em NeurIPS}, 2021.

\bibitem{sohn2016improved}
K.~Sohn,
\newblock ``Improved deep metric learning with multi-class n-pair loss
  objective,''
\newblock in {\em NeurIPS}, 2016.

\bibitem{lee2018stacked}
K.~Lee, X.~Chen, G.~Hua, H.~Hu, and X.~He,
\newblock ``Stacked cross attention for image-text matching,''
\newblock in {\em ECCV}, 2018.

\bibitem{hermans2017defense}
A.~Hermans, L.~Beyer, and B.~Leibe,
\newblock ``In defense of the triplet loss for person re-identification,''
\newblock {\em arXiv preprint arXiv:1703.07737}, 2017.

\bibitem{oh2016deep}
H.~Oh~Song, Y.~Xiang, S.~Jegelka, and S.~Savarese,
\newblock ``Deep metric learning via lifted structured feature embedding,''
\newblock in {\em CVPR}, 2016.

\bibitem{schroff2015facenet}
F.~Schroff, D.~Kalenichenko, and J.~Philbin,
\newblock ``Facenet: A unified embedding for face recognition and clustering,''
\newblock in {\em CVPR}, 2015.

\bibitem{wang2019ranked}
X.~Wang, Y.~Hua, E.~Kodirov, G.~Hu, R.~Garnier, and N.~M. Robertson,
\newblock ``Ranked list loss for deep metric learning,''
\newblock in {\em CVPR}, 2019.

\bibitem{msloss}
X.~Wang, X.~Han, W.~Huang, D.~Dong, and M.~R. Scott,
\newblock ``Multi-similarity loss with general pair weighting for deep metric
  learning,''
\newblock in {\em CVPR}, 2019.

\bibitem{CLIP}
A.~Radford, J.~W. Kim, C.~Hallacy, A.~Ramesh, G.~Goh, S.~Agarwal, G.~Sastry,
  A.~Askell, P.~Mishkin, J.~Clark, et~al.,
\newblock ``Learning transferable visual models from natural language
  supervision,''
\newblock in {\em ICML}, 2021.

\bibitem{cl2021}
Y.~Rao, G.~Chen, J.~Lu, and J.~Zhou,
\newblock ``Counterfactual attention learning for fine-grained visual
  categorization and re-identification,''
\newblock in {\em CVPR}, 2021.

\bibitem{CL}
Z.~Liang, W.~Jiang, H.~Hu, and J.~Zhu,
\newblock ``Learning to contrast the counterfactual samples for robust visual
  question answering,''
\newblock in {\em EMNLP}, 2020.

\bibitem{cl2020vqa}
E.~Abbasnejad, D.~Teney, A.~Parvaneh, J.~Shi, and A.~Hengel,
\newblock ``Counterfactual vision and language learning,''
\newblock in {\em CVPR}, 2020.

\bibitem{relu}
V.~Nair and G.~Hinton,
\newblock ``Rectified linear units improve restricted boltzmann machines,''
\newblock in {\em ICML}, 2010.

\bibitem{gumbel}
E.~Jang, S.~Gu, and B.~Poole,
\newblock ``Categorical reparametrization with gumble-softmax,''
\newblock in {\em ICLR}, 2017.

\bibitem{CSST}
L.~Chen, Y.~Zheng, Y.~Niu, H.~Zhang, and J.~Xiao,
\newblock ``Counterfactual samples synthesizing and training for robust visual
  question answering,''
\newblock {\em arXiv preprint arXiv:2110.01013}, 2021.

\end{thebibliography}

\end{document}